\documentclass[runningheads]{llncs}
\usepackage[T1]{fontenc}
\usepackage{graphicx}
\usepackage{longtable}
\usepackage{multirow}
\usepackage{booktabs}
\usepackage{pdflscape}
\usepackage{makecell}
\usepackage{tcolorbox}
\usepackage{caption} 
\usepackage{subcaption}
\usepackage{longtable}
\usepackage{tikz}
\usepackage{pgfplots}
\pgfplotsset{compat=1.18}
\usepackage{hyperref}

\begin{document}
\title{Machine Learning-Based Prediction of Childhood Stunting in Bangladesh: Fairness and Temporal Robustness Assessment}
\titlerunning{Machine Learning-Based Prediction of Childhood Stunting}

\author{Md Ahshanul Haque\orcidID{0000-0003-3452-8367}\and
Muhammad Ashad Kabir\orcidID{0000-0002-6798-6535}}
\authorrunning{M. A. Haque and M. A. Kabir}
%
\institute{School of Computing, Mathematics and Engineering,\\Charles Sturt University, NSW, Australia\\
\email{\{mdhaque,akabir\}@csu.edu.au}}
\maketitle              
\begin{abstract}
Childhood stunting remains a major public health concern in Bangladesh and reflects long-term growth failure influenced by child, maternal, household, socioeconomic, and health-service factors. This study used nationally representative Bangladesh Demographic and Health Survey data from 2007 to 2022 to develop machine learning models for population-level prediction of childhood stunting and to assess temporal robustness and subgroup fairness. Children aged 0--59 months with complete anthropometric and predictor data were included. Data from the 2007, 2011, and 2014 survey rounds were used for model development, while the 2018 and 2022 rounds were retained as temporal test datasets. Twelve feature-selection approaches were assessed, and the KNN permutation importance-selected predictor set was used for final model evaluation. Eleven machine learning models were evaluated: ten conventional algorithms and one pretrained tabular foundation model, TabPFN. Performance was assessed using balanced accuracy, AUROC, F1-score, Brier score, and expected calibration error. Subgroup fairness was examined by child sex, place of residence, and socioeconomic status. The final analytic sample included 18,844 children, of whom 35.05\% were stunted. In the development hold-out test dataset, TabPFN showed the highest observed balanced accuracy overall at 67.58\%, while AdaBoost showed the highest observed balanced accuracy among conventional models at 67.51\%. In temporal testing, the highest observed balanced accuracy was found for Gradient Boosting in BDHS 2018 and XGBoost in BDHS 2022. Model performance varied across survey rounds and subgroups, highlighting the importance of temporal validation, subgroup fairness assessment, and transparent interpretation in public health prediction modeling.

\end{abstract}

\section{Introduction}
Childhood malnutrition remains a major public health concern in many low- and middle-income countries. Stunting, wasting, and underweight are widely used indicators of poor nutritional status among children under five. Among these, stunting reflects long-term growth failure and is associated with increased illness, poor growth, and impaired cognitive and physical development \cite{fenta2020,unicef2023}. The burden remains particularly high in South Asia \cite{unicef2018,unicef2025,wali2020}. Although Bangladesh has made substantial progress in reducing undernutrition, stunting remains a continuing concern. According to the Bangladesh Demographic and Health Survey (BDHS) 2022, 24\% of children under five were stunted, 11.7\% were wasted, and 22\% were underweight \cite{niport2024}. Reducing childhood stunting is also central to Sustainable Development Goal (SDG) 2.2, under which Bangladesh aims to reduce under-five stunting to 15.5\% by 2030 \cite{unicef2024}. Achieving this target will require better identification of children at high risk and more targeted public health action \cite{haque2023}.

Childhood stunting is influenced by child, maternal, household, and community-level factors, many of which are routinely collected in Demographic and Health Surveys (DHS) \cite{dhsprogram2026}. Repeated BDHS rounds therefore provide an opportunity to examine both stunting patterns and the stability of prediction models across survey periods. Previous studies in Bangladesh have mainly used conventional statistical models to identify associated factors \cite{hossain2024,islam2020}. However, these models are primarily designed to estimate associations rather than to develop prediction models, which require explicit evaluation of predictive performance, calibration, validation, and generalizability. Although machine learning is increasingly used for population-level prediction, existing studies in Bangladesh have mostly relied on single datasets and within-sample validation, limiting evidence on temporal robustness \cite{islam2024,mansur2021,rahman2021,talukder2020,tamanna2025}. To address this gap, this study used repeated nationally representative BDHS data from 2007 to 2022 to develop and evaluate machine learning models for childhood stunting. The study assessed predictive performance using a development hold-out test dataset and temporal test datasets, identified prediction-relevant predictors across multiple epidemiological domains, and examined subgroup performance by sex, residence, and socioeconomic status.

\section{Related Work}
Research on childhood stunting and undernutrition in Bangladesh has grown over the last two decades, with most studies using cross-sectional secondary data, particularly BDHS \cite{das2019,hossain2020,islam2025,khan2024,kundu2022,rahman2020,tamanna2025}. These studies have identified a wide range of associated factors, including child age, birth weight, recent illness, feeding practices, maternal education and nutritional status, antenatal care, household wealth, sanitation, drinking water source, food insecurity, residence, and administrative division \cite{alom2012,anik2021,choudhury2017,chowdhury2016,haque2021a,talukder2020,tamanna2025}. Although this evidence is important, most studies were designed to estimate associations rather than develop prediction models or identify predictors most useful for classification.

Machine learning has been used less often for childhood stunting research in Bangladesh. Existing studies have applied models such as logistic regression, decision tree, random forest, support vector machine, k-nearest neighbors, neural networks, and gradient boosting to predict or classify child malnutrition outcomes \cite{islam2024,mansur2021,rahman2021,talukder2020,tamanna2025}. Some studies also reported variable importance, supporting knowledge discovery. However, most relied on a single dataset and within-sample validation, such as random train-test splits or cross-validation within the same survey round, limiting evidence on temporal robustness.

Machine learning studies from other low- and middle-income countries have also used DHS or MICS data to predict childhood undernutrition \cite{anku2024,bitew2022,chilyabanyama2022,fenta2021,shen2023}. Studies from Ethiopia, Zambia, Papua New Guinea, and Ghana reported good performance from Random Forest, XGBoost, and related ensemble methods. However, most studies were based on a single survey round, with limited assessment of performance over time. This gap is important because models that perform well in one survey may not remain stable in later rounds. Repeated national survey data are therefore needed to assess temporal robustness, subgroup performance, and knowledge discovery across survey periods.

\section{Methods}

\begin{figure}[!t]
\centering
\includegraphics[width=\textwidth]{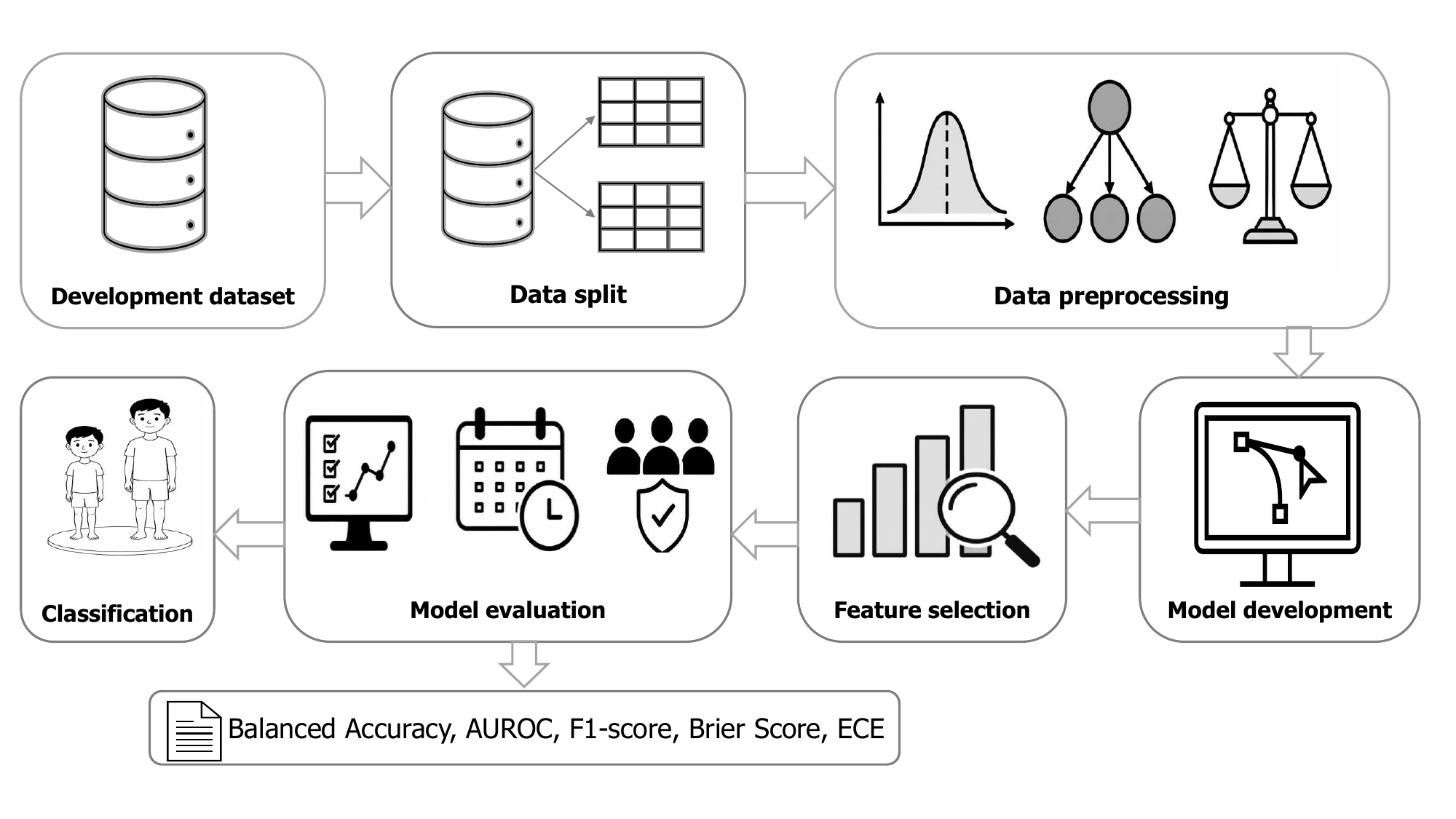}
\caption{Overview of the machine learning workflow for childhood stunting prediction using repeated BDHS data.}
\label{fig:workflow}
\end{figure}

Fig.~\ref{fig:workflow} presents the overall methodological workflow of the study. BDHS 2007, 2011, and 2014 were combined for model development, while BDHS 2018 and 2022 were reserved for temporal testing. After data preparation, feature selection was performed using 12 approaches, followed by model development using 10 conventional machine learning algorithms and the pretrained tabular foundation model, TabPFN. Model performance was evaluated using a development hold-out test dataset, temporal test datasets, and subgroup fairness assessment. Prediction-relevant predictors were identified through the feature selection process to support knowledge discovery.

\subsection{Datasets}

This study used repeated nationally representative BDHS data \cite{dhsprogram2026}. The study population included children aged 0--59 months with available anthropometric information from the selected survey rounds. The participant selection process and stunting distribution by survey year are shown in Table~\ref{tab:participant_selection}. Across the BDHS 2007, 2011, 2014, 2018, and 2022 survey rounds, 40,332 children were initially eligible. Information availability varied across participants because some variables, including anthropometric measurements and selected maternal health-care characteristics, were not collected for all participants in the original surveys. Children without the anthropometric information required to determine stunting status were therefore excluded. Missing outcome values were not imputed or otherwise estimated because the stunting outcome could not be established from unobserved anthropometric measurements. Children with missing information on at least one key predictor were also excluded to maintain a consistent predictor set across the analysis. The final analytic sample included 18,844 children. In the final analytic sample, 6,605 children were stunted, giving an overall stunting prevalence of 35.05\%, while 12,239 children were non-stunted. Stunting prevalence decreased across survey years, from 40.23\% in 2007 and 40.05\% in 2011 to 32.61\% in 2014, 31.05\% in 2018, and 22.82\% in 2022. Both descriptive and machine learning analyses were conducted using this final complete-case sample.

\begin{table}[!t]
\caption{Participant selection by BDHS survey year. Values for stunted and non-stunted children are presented as $n$ (\%).}
\label{tab:participant_selection}
\centering
\begin{tabular*}{\textwidth}{@{\extracolsep{\fill}}lcccc@{}}
\toprule
Year & Survey & Selected & Stunted & Non-stunted \\
     &  sample  & sample ($N$) & \textit{n} (\%) & \textit{n} (\%) \\
\midrule
2007  & 6,150  & 3,813 & 1,534 (40.23) & 2,279 (59.77) \\
2011  & 8,753  & 5,673 & 2,272 (40.05) & 3,401 (59.95) \\
2014  & 7,886  & 3,527 & 1,150 (32.61) & 2,377 (67.39) \\
2018  & 8,759  & 3,868 & 1,201 (31.05) & 2,667 (68.95) \\
2022  & 8,784  & 1,963 &   448 (22.82) & 1,515 (77.18) \\
\midrule
Total & 40,332 & 18,844 & 6,605 (35.05) & 12,239 (64.95) \\
\bottomrule
\end{tabular*}
\end{table}

The feature summary is presented in Table~\ref{tab:characteristics}. Overall, the distribution of children across wealth index categories was relatively balanced: 21.2\% were from the poorest households, 19.6\% from poorer households, 18.8\% from the middle wealth category, 19.7\% from richer households, and 20.6\% from the richest households. About 39.5\% of children lived in households with a hygienic toilet, and 51.9\% were male. The mean age of children was 22.5 months (standard deviation [SD]: 14.9 months). Regarding feeding and health-related characteristics, 75.3\% of children were currently breastfed, and 71.9\% had early initiation of breastfeeding. Fever in the last two weeks was reported for 38.4\% of children, and 64.0\% had received vitamin A supplementation in the previous six months. Overall, 28.7\% of mothers had received at least four antenatal care visits from a trained provider.

For parental and household characteristics, 76.1\% of fathers had formal education, and 23.9\% were engaged in agriculture. The mean maternal height was 151.0 cm (SD: 5.7 cm), and the mean maternal age at first birth was 18.3 years (SD: 3.3 years). About 24.8\% of mothers had worked in the last 12 months. Non-caesarean delivery was reported for 77.8\% of births. In addition, 12.8\% of children had at least one sibling who had died, 59.8\% had a birth interval of 24 months or longer, 62.2\% had exposure to mass media, and 26.8\% were from households where attitudes supportive of wife beating were reported.

\begin{table}[!t]
\caption{Characteristics of children by stunting status in the pooled BDHS analytic sample. Categorical variables are presented as $n$ (\%); continuous variables are presented as mean $\pm$ SD.}
\label{tab:characteristics}
\centering
\resizebox{\textwidth}{!}{
\begin{tabular}{llrrrr}
\toprule
Feature & Value & Non-stunted & Stunted & Overall & $p$-value \\
& & ($N=12,239$) & ($N=6,605$) & ($N=18,844$) & \\
\midrule

Wealth index & Poorest & 2,121 (17.3) & 1,883 (28.5) & 4,004 (21.2) & $<0.001$ \\
& Poorer & 2,214 (18.1) & 1,481 (22.4) & 3,695 (19.6) & \\
& Middle & 2,287 (18.7) & 1,264 (19.1) & 3,551 (18.8) & \\
& Richer & 2,588 (21.1) & 1,132 (17.1) & 3,720 (19.7) & \\
& Richest & 3,029 (24.7) & 845 (12.8) & 3,874 (20.6) & \\

Hygienic toilet & Yes & 5,418 (44.3) & 2,021 (30.6) & 7,439 (39.5) & $<0.001$ \\

Child's sex & Male & 6,251 (51.1) & 3,521 (53.3) & 9,772 (51.9) & 0.004 \\

Child's age (months) & Number & 20.5 $\pm$ 14.9 & 26.2 $\pm$ 14.0 & 22.5 $\pm$ 14.9 & $<0.001$ \\

Currently breastfed & Yes & 9,413 (76.9) & 4,768 (72.2) & 14,181 (75.3) & $<0.001$ \\

Early initiation of breastfeeding & Yes & 8,689 (71.0) & 4,860 (73.6) & 13,549 (71.9) & $<0.001$ \\

Fever in the last 2 weeks & Yes & 4,538 (37.1) & 2,695 (40.8) & 7,233 (38.4) & $<0.001$ \\

Father education $\geq1$ year & Yes & 9,812 (80.2) & 4,522 (68.5) & 14,334 (76.1) & $<0.001$ \\

Father's occupation & Agriculture & 2,609 (21.3) & 1,904 (28.8) & 4,513 (23.9) & $<0.001$ \\ & Worker or laborer & 5,507 (45.0) & 3,100 (46.9) & 8,607 (45.7) & \\ & Business or professional & 3,811 (31.1) & 1,452 (22.0) & 5,263 (27.9) & \\ & Not working & 312 (2.5) & 149 (2.3) & 461 (2.4) & \\  

Mother's height (cm) & Number & 151.9 $\pm$ 5.6 & 149.2 $\pm$ 5.5 & 151.0 $\pm$ 5.7 & $<0.001$ \\

Mother worked in the last 12 months & Yes & 2,966 (24.2) & 1,709 (25.9) & 4,675 (24.8) & 0.014 \\

Vitamin A supplementation & Yes & 7,529 (61.5) & 4,528 (68.6) & 12,057 (64.0) & $<0.001$ \\

At least four ANC visits & Yes & 4,058 (33.2) & 1,359 (20.6) & 5,417 (28.7) & $<0.001$ \\

Non-caesarean delivery & Yes & 8,999 (73.5) & 5,666 (85.8) & 14,665 (77.8) & $<0.001$ \\

Mother's age at first birth & Number & 18.5 $\pm$ 3.4 & 17.8 $\pm$ 3.0 & 18.3 $\pm$ 3.3 & $<0.001$ \\

At least one child died & Yes & 1,362 (11.1) & 1,048 (15.9) & 2,410 (12.8) & $<0.001$ \\

Birth interval $\geq24$ months & Yes & 7,221 (59.0) & 4,046 (61.3) & 11,267 (59.8) & 0.003 \\

Mass media exposure & Yes & 8,062 (65.9) & 3,664 (55.5) & 11,726 (62.2) & $<0.001$ \\

Wife beating justified & Yes & 3,014 (24.6) & 2,044 (30.9) & 5,058 (26.8) & $<0.001$ \\

\bottomrule
\end{tabular}
}
\end{table}

\subsection{Train and Test Splits}

Data from the 2007, 2011, and 2014 BDHS rounds were combined for model development. This pooled dataset included 13,013 children and was split into 80\% training data and a 20\% development hold-out test dataset. Stratified sampling was applied using the outcome and key variables, including survey year, socioeconomic status, residence, and child sex. The final training dataset included 10,410 children, and the development hold-out test dataset included 2,603 children. The 2018 and 2022 BDHS rounds, including 3,868 and 1,963 children, respectively, were retained as temporally distinct test datasets to evaluate model generalizability over time.

\subsection{Data Preprocessing}

Variables were prepared in numeric format before model development. Binary variables were coded as 0 or 1, and categorical variables were converted using one-hot encoding. Continuous variables were scaled using \textit{StandardScaler}, with the mean and standard deviation estimated from the training dataset only. The same scaling parameters were then applied to the development hold-out and temporal test datasets to avoid data leakage. 

\subsection{Feature Selection}

Feature selection was performed using the training dataset only. Twelve feature-selection approaches were applied: recursive feature elimination with cross-validation (RFECV) using eight estimators—Logistic Regression, Decision Tree, Support Vector Machine, Random Forest, Extra Trees, AdaBoost, Gradient Boosting, and XGBoost; permutation importance using KNN and MLP; the Mann–Whitney U test; and LASSO-regularized Logistic Regression \cite{kabir2026}. Feature selection was used not only to reduce dimensionality, but also to support knowledge discovery by identifying prediction-relevant variables for childhood stunting. The predictive performance of the resulting feature-selection sets was compared using the development hold-out test dataset. The KNN permutation importance-selected feature set showed the highest observed balanced accuracy among the evaluated feature-selection sets and was therefore carried forward for subsequent model evaluation and stratified performance assessment.

\subsection{Model Development}

Eleven machine learning models were evaluated: ten conventional machine learning classifiers and one pretrained tabular foundation model, TabPFN. The ten conventional classifiers were Logistic Regression (LR), Decision Tree (DT), K-Nearest Neighbors (KNN), Support Vector Machine (SVM), Random Forest (RF), Extra Trees (ET), AdaBoost, Gradient Boosting (GB), XGBoost, and Multilayer Perceptron (MLP).

The 80\% training data from the pooled BDHS 2007, 2011, and 2014 rounds were used for model development, including feature selection, class balancing, hyperparameter tuning, and final model training. To address class imbalance during model training, random undersampling was applied only to the training portion of each cross-validation fold. The validation portion of each fold was retained with its original class distribution and was not used for undersampling. Thus, undersampling was performed independently within each training fold and did not involve the validation data. No undersampling was applied to the development hold-out or temporal test datasets. Hyperparameter tuning was conducted using \textit{Optuna} on the training dataset only. Model performance during tuning was assessed using 10-fold cross-validation, with random undersampling performed separately within the training portion of each fold. Balanced accuracy was used as the primary optimization metric because the outcome was imbalanced. The best hyperparameters obtained during training-stage tuning were then used to train the final conventional machine learning models using the KNN permutation importance-selected predictors. For the final training step, class balancing was performed using random undersampling of the training data. For TabPFN, the same KNN permutation importance-selected predictors were used. Although TabPFN uses pretrained model weights, the labeled training data were provided as context for generating predictions. This allowed TabPFN to be evaluated using the same development hold-out and temporal test datasets as the conventional machine learning models.

Final model performance was assessed on the development hold-out test dataset derived from BDHS 2007, 2011, and 2014, and on the BDHS 2018 and BDHS 2022 temporal test datasets. No resampling was performed in any of these evaluation datasets, so their original class distributions were retained. The main evaluation metrics were balanced accuracy, area under the receiver operating characteristic curve, F1-score, Brier score, and expected calibration error. For overall model performance, BCa 95\% confidence intervals were estimated using bias-corrected and accelerated bootstrap resampling. Bootstrap resampling was applied to child-level prediction results within each test dataset, and performance metrics were recalculated across bootstrap samples.

\section{Results}

\subsection{Selected Predictors}

Twelve feature-selection approaches were compared to identify predictors relevant for childhood stunting prediction, including Recursive Feature Elimination with Cross-Validation (RFECV), permutation importance, Mann--Whitney U test, and LASSO Logistic Regression. The KNN permutation importance-selected feature set showed the highest observed balanced accuracy in the development hold-out test dataset and was therefore used for final model development and evaluation. The selected predictors covered multiple epidemiological domains, including child age and sex, breastfeeding status, early initiation of breastfeeding, recent fever, vitamin A supplementation, birth interval, history of child death, maternal height, maternal age at first birth, maternal working status, antenatal care, delivery mode, household socioeconomic status, hygienic toilet, father's education and occupation, mass media exposure, and attitudes toward wife beating. These variables were considered prediction-relevant predictors and should not be interpreted as causal determinants of childhood stunting.

\definecolor{ash}{gray}{0.4}
\small
\begin{longtable}[c]{llccccc}
\caption{Overall model performance across the development hold-out test dataset derived from BDHS 2007, 2011, and 2014, and the BDHS 2018 and 2022 temporal test datasets. Bold values indicate the highest observed value within each test dataset and metric for balanced accuracy, AUROC, and F1, and the lowest observed value for Brier score and ECE. Values are presented as point estimates with BCa 95\% confidence intervals in parentheses.}\label{tab:model_performance}\\


\toprule
Test set & Model & 
B. Accuracy ($\uparrow$) & 
AUROC ($\uparrow$) & 
F1 ($\uparrow$) & 
Brier ($\downarrow$) & 
ECE ($\downarrow$) \\
\midrule
\endfirsthead

\multicolumn{7}{l}{\textit{\tablename\ \thetable{} -- continued from previous page}} \\
\toprule
Test set & Model & 
B. Accuracy ($\uparrow$) & 
AUROC ($\uparrow$) & 
F1 ($\uparrow$) & 
Brier ($\downarrow$) & 
ECE ($\downarrow$) \\
\midrule
\endhead

\midrule
\multicolumn{7}{r}{\textit{continued on next page}} 
\endfoot

\bottomrule
\endlastfoot


\multirow{18}{*}{\rotatebox[origin=c]{90}{ 2007--2014 20\% hold-out}}
& LR & \makecell[t]{65.59\\[-4pt]{\scriptsize\textcolor{ash}{[63.63, 67.55]}}} & \makecell[t]{0.707\\[-4pt]{\scriptsize\textcolor{ash}{[0.685, 0.727]}}} & \makecell[t]{64.35\\[-4pt]{\scriptsize\textcolor{ash}{[62.37, 66.28]}}} & \makecell[t]{0.219\\[-4pt]{\scriptsize\textcolor{ash}{[0.213, 0.226]}}} & \makecell[t]{0.269\\[-4pt]{\scriptsize\textcolor{ash}{[0.250, 0.287]}}} \\
& DT & \makecell[t]{64.83\\[-4pt]{\scriptsize\textcolor{ash}{[62.85, 66.63]}}} & \makecell[t]{0.704\\[-4pt]{\scriptsize\textcolor{ash}{[0.683, 0.725]}}} & \makecell[t]{62.39\\[-4pt]{\scriptsize\textcolor{ash}{[60.38, 64.29]}}} & \makecell[t]{0.222\\[-4pt]{\scriptsize\textcolor{ash}{[0.215, 0.228]}}} & \makecell[t]{0.291\\[-4pt]{\scriptsize\textcolor{ash}{[0.273, 0.310]}}} \\
& KNN & \makecell[t]{64.34\\[-4pt]{\scriptsize\textcolor{ash}{[62.20, 66.20]}}} & \makecell[t]{0.701\\[-4pt]{\scriptsize\textcolor{ash}{[0.680, 0.723]}}} & \makecell[t]{63.19\\[-4pt]{\scriptsize\textcolor{ash}{[61.14, 64.98]}}} & \makecell[t]{0.219\\[-4pt]{\scriptsize\textcolor{ash}{[0.213, 0.226]}}} & \makecell[t]{0.260\\[-4pt]{\scriptsize\textcolor{ash}{[0.242, 0.279]}}} \\
& SVM & \makecell[t]{66.49\\[-4pt]{\scriptsize\textcolor{ash}{[64.55, 68.31]}}} & \makecell[t]{0.726\\[-4pt]{\scriptsize\textcolor{ash}{[0.706, 0.747]}}} & \makecell[t]{65.26\\[-4pt]{\scriptsize\textcolor{ash}{[63.29, 67.15]}}} & \makecell[t]{0.213\\[-4pt]{\scriptsize\textcolor{ash}{[0.206, 0.220]}}} & \makecell[t]{0.292\\[-4pt]{\scriptsize\textcolor{ash}{[0.274, 0.311]}}} \\
& RF & \makecell[t]{67.08\\[-4pt]{\scriptsize\textcolor{ash}{[65.23, 69.03]}}} & \makecell[t]{0.728\\[-4pt]{\scriptsize\textcolor{ash}{[0.707, 0.748]}}} & \makecell[t]{65.40\\[-4pt]{\scriptsize\textcolor{ash}{[63.46, 67.21]}}} & \makecell[t]{0.213\\[-4pt]{\scriptsize\textcolor{ash}{[0.208, 0.219]}}} & \makecell[t]{0.263\\[-4pt]{\scriptsize\textcolor{ash}{[0.245, 0.282]}}} \\

& ET & \makecell[t]{66.89\\[-4pt]{\scriptsize\textcolor{ash}{[64.99, 68.84]}}} & \makecell[t]{0.720\\[-4pt]{\scriptsize\textcolor{ash}{[0.699, 0.741]}}} & \makecell[t]{65.05\\[-4pt]{\scriptsize\textcolor{ash}{[63.14, 67.02]}}} & \makecell[t]{0.216\\[-4pt]{\scriptsize\textcolor{ash}{[0.210, 0.223]}}} & \makecell[t]{0.273\\[-4pt]{\scriptsize\textcolor{ash}{[0.255, 0.292]}}} \\

& AdaBoost & \makecell[t]{67.51\\[-4pt]{\scriptsize\textcolor{ash}{[65.59, 69.39]}}} & \makecell[t]{0.730\\[-4pt]{\scriptsize\textcolor{ash}{[0.708, 0.748]}}} & \textbf{\makecell[t]{66.01\\[-4pt]{\scriptsize\textcolor{ash}{[64.19, 67.86]}}}} & \makecell[t]{0.219\\[-4pt]{\scriptsize\textcolor{ash}{[0.215, 0.223]}}} & \textbf{\makecell[t]{0.217\\[-4pt]{\scriptsize\textcolor{ash}{[0.197, 0.233]}}}} \\

& GB & \makecell[t]{67.06\\[-4pt]{\scriptsize\textcolor{ash}{[65.11, 68.91]}}} & \makecell[t]{0.729\\[-4pt]{\scriptsize\textcolor{ash}{[0.708, 0.748]}}} & \makecell[t]{65.56\\[-4pt]{\scriptsize\textcolor{ash}{[63.59, 67.34]}}} & \makecell[t]{0.213\\[-4pt]{\scriptsize\textcolor{ash}{[0.207, 0.219]}}} & \makecell[t]{0.274\\[-4pt]{\scriptsize\textcolor{ash}{[0.256, 0.293]}}} \\
& XGBoost & \makecell[t]{66.75\\[-4pt]{\scriptsize\textcolor{ash}{[64.87, 68.70]}}} & \makecell[t]{0.729\\[-4pt]{\scriptsize\textcolor{ash}{[0.708, 0.748]}}} & \makecell[t]{65.12\\[-4pt]{\scriptsize\textcolor{ash}{[63.23, 66.97]}}} & \textbf{\makecell[t]{0.212\\[-4pt]{\scriptsize\textcolor{ash}{[0.206, 0.219]}}}} & \makecell[t]{0.293\\[-4pt]{\scriptsize\textcolor{ash}{[0.275, 0.313]}}} \\
& MLP & \makecell[t]{66.41\\[-4pt]{\scriptsize\textcolor{ash}{[64.53, 68.32]}}} & \makecell[t]{0.722\\[-4pt]{\scriptsize\textcolor{ash}{[0.699, 0.741]}}} & \makecell[t]{65.53\\[-4pt]{\scriptsize\textcolor{ash}{[63.62, 67.41]}}} & \textbf{\makecell[t]{0.212\\[-4pt]{\scriptsize\textcolor{ash}{[0.206, 0.219]}}}} & \makecell[t]{0.273\\[-4pt]{\scriptsize\textcolor{ash}{[0.254, 0.291]}}} \\
& TabPFN & \textbf{\makecell[t]{67.58\\[-4pt]{\scriptsize\textcolor{ash}{[65.68, 69.50]}}}} & \textbf{\makecell[t]{0.734\\[-4pt]{\scriptsize\textcolor{ash}{[0.713, 0.753]}}}} & \makecell[t]{65.98\\[-4pt]{\scriptsize\textcolor{ash}{[64.11, 67.89]}}} & \textbf{\makecell[t]{0.212\\[-4pt]{\scriptsize\textcolor{ash}{[0.204, 0.219]}}}} & \makecell[t]{0.310\\[-4pt]{\scriptsize\textcolor{ash}{[0.292, 0.330]}}} \\

\midrule

\multirow{18}{*}{\rotatebox[origin=c]{90}{BDHS 2018}}
& LR & \makecell[t]{63.76\\[-4pt]{\scriptsize\textcolor{ash}{[62.11, 65.37]}}} & \makecell[t]{0.696\\[-4pt]{\scriptsize\textcolor{ash}{[0.679, 0.714]}}} & \makecell[t]{63.65\\[-4pt]{\scriptsize\textcolor{ash}{[62.05, 65.24]}}} & \makecell[t]{0.204\\[-4pt]{\scriptsize\textcolor{ash}{[0.199, 0.208]}}} & \makecell[t]{0.349\\[-4pt]{\scriptsize\textcolor{ash}{[0.335, 0.365]}}} \\
& DT & \makecell[t]{64.22\\[-4pt]{\scriptsize\textcolor{ash}{[62.72, 65.97]}}} & \makecell[t]{0.692\\[-4pt]{\scriptsize\textcolor{ash}{[0.674, 0.710]}}} & \makecell[t]{62.01\\[-4pt]{\scriptsize\textcolor{ash}{[60.59, 63.80]}}} & \makecell[t]{0.210\\[-4pt]{\scriptsize\textcolor{ash}{[0.203, 0.215]}}} & \makecell[t]{0.380\\[-4pt]{\scriptsize\textcolor{ash}{[0.366, 0.396]}}} \\
& KNN & \makecell[t]{64.55\\[-4pt]{\scriptsize\textcolor{ash}{[62.95, 66.26]}}} & \makecell[t]{0.697\\[-4pt]{\scriptsize\textcolor{ash}{[0.679, 0.715]}}} & \makecell[t]{63.81\\[-4pt]{\scriptsize\textcolor{ash}{[62.27, 65.46]}}} & \makecell[t]{0.209\\[-4pt]{\scriptsize\textcolor{ash}{[0.204, 0.213]}}} & \makecell[t]{0.336\\[-4pt]{\scriptsize\textcolor{ash}{[0.323, 0.352]}}} \\
& SVM & \makecell[t]{64.94\\[-4pt]{\scriptsize\textcolor{ash}{[63.34, 66.59]}}} & \makecell[t]{0.713\\[-4pt]{\scriptsize\textcolor{ash}{[0.696, 0.731]}}} & \makecell[t]{64.58\\[-4pt]{\scriptsize\textcolor{ash}{[63.04, 66.20]}}} & \textbf{\makecell[t]{0.200\\[-4pt]{\scriptsize\textcolor{ash}{[0.194, 0.205]}}}} & \makecell[t]{0.368\\[-4pt]{\scriptsize\textcolor{ash}{[0.354, 0.384]}}} \\
& RF & \makecell[t]{65.25\\[-4pt]{\scriptsize\textcolor{ash}{[63.71, 66.84]}}} & \makecell[t]{0.713\\[-4pt]{\scriptsize\textcolor{ash}{[0.696, 0.732]}}} & \makecell[t]{64.37\\[-4pt]{\scriptsize\textcolor{ash}{[62.95, 66.01]}}} & \makecell[t]{0.203\\[-4pt]{\scriptsize\textcolor{ash}{[0.198, 0.208]}}} & \makecell[t]{0.344\\[-4pt]{\scriptsize\textcolor{ash}{[0.330, 0.360]}}} \\
& ET & \makecell[t]{64.53\\[-4pt]{\scriptsize\textcolor{ash}{[62.82, 66.13]}}} & \makecell[t]{0.703\\[-4pt]{\scriptsize\textcolor{ash}{[0.685, 0.721]}}} & \makecell[t]{63.85\\[-4pt]{\scriptsize\textcolor{ash}{[62.24, 65.43]}}} & \makecell[t]{0.204\\[-4pt]{\scriptsize\textcolor{ash}{[0.199, 0.209]}}} & \makecell[t]{0.355\\[-4pt]{\scriptsize\textcolor{ash}{[0.341, 0.372]}}} \\
& AdaBoost & \makecell[t]{65.19\\[-4pt]{\scriptsize\textcolor{ash}{[63.67, 66.87]}}} & \textbf{\makecell[t]{0.714\\[-4pt]{\scriptsize\textcolor{ash}{[0.697, 0.731]}}}} & \makecell[t]{64.43\\[-4pt]{\scriptsize\textcolor{ash}{[62.94, 66.03]}}} & \makecell[t]{0.210\\[-4pt]{\scriptsize\textcolor{ash}{[0.206, 0.213]}}} & \textbf{\makecell[t]{0.299\\[-4pt]{\scriptsize\textcolor{ash}{[0.285, 0.313]}}}} \\
& GB & \textbf{\makecell[t]{65.56\\[-4pt]{\scriptsize\textcolor{ash}{[63.90, 67.05]}}}} & \textbf{\makecell[t]{0.714\\[-4pt]{\scriptsize\textcolor{ash}{[0.697, 0.733]}}}} & \textbf{\makecell[t]{64.77\\[-4pt]{\scriptsize\textcolor{ash}{[63.22, 66.27]}}}} & \makecell[t]{0.201\\[-4pt]{\scriptsize\textcolor{ash}{[0.196, 0.206]}}} & \makecell[t]{0.360\\[-4pt]{\scriptsize\textcolor{ash}{[0.346, 0.376]}}} \\
& XGBoost & \makecell[t]{65.24\\[-4pt]{\scriptsize\textcolor{ash}{[63.64, 66.96]}}} & \textbf{\makecell[t]{0.714\\[-4pt]{\scriptsize\textcolor{ash}{[0.697, 0.733]}}}} & \makecell[t]{64.32\\[-4pt]{\scriptsize\textcolor{ash}{[62.76, 65.96]}}} & \makecell[t]{0.201\\[-4pt]{\scriptsize\textcolor{ash}{[0.196, 0.207]}}} & \makecell[t]{0.378\\[-4pt]{\scriptsize\textcolor{ash}{[0.364, 0.395]}}} \\
& MLP & \makecell[t]{63.71\\[-4pt]{\scriptsize\textcolor{ash}{[62.15, 65.26]}}} & \makecell[t]{0.704\\[-4pt]{\scriptsize\textcolor{ash}{[0.687, 0.722]}}} & \makecell[t]{63.84\\[-4pt]{\scriptsize\textcolor{ash}{[62.29, 65.42]}}} & \textbf{\makecell[t]{0.200\\[-4pt]{\scriptsize\textcolor{ash}{[0.195, 0.204]}}}} & \makecell[t]{0.354\\[-4pt]{\scriptsize\textcolor{ash}{[0.340, 0.370]}}} \\
& TabPFN & \makecell[t]{64.85\\[-4pt]{\scriptsize\textcolor{ash}{[63.28, 66.43]}}} & \textbf{\makecell[t]{0.714\\[-4pt]{\scriptsize\textcolor{ash}{[0.697, 0.733]}}}} & \makecell[t]{64.17\\[-4pt]{\scriptsize\textcolor{ash}{[62.64, 65.78]}}} & \makecell[t]{0.201\\[-4pt]{\scriptsize\textcolor{ash}{[0.194, 0.207]}}} & \makecell[t]{0.396\\[-4pt]{\scriptsize\textcolor{ash}{[0.382, 0.413]}}} \\

\midrule

& LR & \makecell[t]{62.98\\[-4pt]{\scriptsize\textcolor{ash}{[60.49, 65.41]}}} & \makecell[t]{0.689\\[-4pt]{\scriptsize\textcolor{ash}{[0.660, 0.718]}}} & \makecell[t]{62.85\\[-4pt]{\scriptsize\textcolor{ash}{[60.44, 65.27]}}} & \makecell[t]{0.185\\[-4pt]{\scriptsize\textcolor{ash}{[0.179, 0.193]}}} & \makecell[t]{0.448\\[-4pt]{\scriptsize\textcolor{ash}{[0.427, 0.466]}}} \\
& DT & \makecell[t]{63.87\\[-4pt]{\scriptsize\textcolor{ash}{[61.43, 66.47]}}} & \makecell[t]{0.686\\[-4pt]{\scriptsize\textcolor{ash}{[0.658, 0.713]}}} & \makecell[t]{59.25\\[-4pt]{\scriptsize\textcolor{ash}{[57.00, 61.65]}}} & \makecell[t]{0.200\\[-4pt]{\scriptsize\textcolor{ash}{[0.193, 0.208]}}} & \makecell[t]{0.464\\[-4pt]{\scriptsize\textcolor{ash}{[0.444, 0.483]}}} \\
\multirow{14}{*}{\rotatebox[origin=c]{90}{ BDHS 2022}} & KNN & \makecell[t]{63.01\\[-4pt]{\scriptsize\textcolor{ash}{[60.17, 65.61]}}} & \makecell[t]{0.686\\[-4pt]{\scriptsize\textcolor{ash}{[0.654, 0.716]}}} & \makecell[t]{62.04\\[-4pt]{\scriptsize\textcolor{ash}{[59.56, 64.58]}}} & \makecell[t]{0.195\\[-4pt]{\scriptsize\textcolor{ash}{[0.189, 0.202]}}} & \makecell[t]{0.429\\[-4pt]{\scriptsize\textcolor{ash}{[0.409, 0.447]}}} \\
& SVM & \makecell[t]{63.52\\[-4pt]{\scriptsize\textcolor{ash}{[61.11, 66.20]}}} & \makecell[t]{0.702\\[-4pt]{\scriptsize\textcolor{ash}{[0.673, 0.731]}}} & \makecell[t]{62.78\\[-4pt]{\scriptsize\textcolor{ash}{[60.51, 65.29]}}} & \makecell[t]{0.185\\[-4pt]{\scriptsize\textcolor{ash}{[0.178, 0.193]}}} & \makecell[t]{0.461\\[-4pt]{\scriptsize\textcolor{ash}{[0.441, 0.480]}}} \\
& RF & \makecell[t]{63.74\\[-4pt]{\scriptsize\textcolor{ash}{[61.09, 66.31]}}} & \makecell[t]{0.704\\[-4pt]{\scriptsize\textcolor{ash}{[0.676, 0.733]}}} & \makecell[t]{61.87\\[-4pt]{\scriptsize\textcolor{ash}{[59.52, 64.33]}}} & \makecell[t]{0.192\\[-4pt]{\scriptsize\textcolor{ash}{[0.185, 0.199]}}} & \makecell[t]{0.435\\[-4pt]{\scriptsize\textcolor{ash}{[0.415, 0.453]}}} \\
& ET & \makecell[t]{63.05\\[-4pt]{\scriptsize\textcolor{ash}{[60.45, 65.69]}}} & \makecell[t]{0.698\\[-4pt]{\scriptsize\textcolor{ash}{[0.669, 0.726]}}} & \makecell[t]{61.53\\[-4pt]{\scriptsize\textcolor{ash}{[59.24, 63.96]}}} & \makecell[t]{0.193\\[-4pt]{\scriptsize\textcolor{ash}{[0.186, 0.200]}}} & \makecell[t]{0.443\\[-4pt]{\scriptsize\textcolor{ash}{[0.423, 0.462]}}} \\
& AdaBoost & \makecell[t]{64.25\\[-4pt]{\scriptsize\textcolor{ash}{[61.76, 66.92]}}} & \textbf{\makecell[t]{0.710\\[-4pt]{\scriptsize\textcolor{ash}{[0.684, 0.740]}}}} & \makecell[t]{62.47\\[-4pt]{\scriptsize\textcolor{ash}{[60.21, 64.83]}}} & \makecell[t]{0.198\\[-4pt]{\scriptsize\textcolor{ash}{[0.194, 0.203]}}} & \textbf{\makecell[t]{0.392\\[-4pt]{\scriptsize\textcolor{ash}{[0.372, 0.409]}}}} \\
& GB & \makecell[t]{63.79\\[-4pt]{\scriptsize\textcolor{ash}{[61.43, 66.42]}}} & \makecell[t]{0.707\\[-4pt]{\scriptsize\textcolor{ash}{[0.680, 0.736]}}} & \makecell[t]{62.00\\[-4pt]{\scriptsize\textcolor{ash}{[59.90, 64.50]}}} & \makecell[t]{0.189\\[-4pt]{\scriptsize\textcolor{ash}{[0.182, 0.197]}}} & \makecell[t]{0.453\\[-4pt]{\scriptsize\textcolor{ash}{[0.433, 0.472]}}} \\
& XGBoost & \textbf{\makecell[t]{64.63\\[-4pt]{\scriptsize\textcolor{ash}{[62.29, 67.33]}}}} & \makecell[t]{0.709\\[-4pt]{\scriptsize\textcolor{ash}{[0.682, 0.738]}}} & \makecell[t]{62.57\\[-4pt]{\scriptsize\textcolor{ash}{[60.37, 65.03]}}} & \makecell[t]{0.188\\[-4pt]{\scriptsize\textcolor{ash}{[0.180, 0.196]}}} & \makecell[t]{0.470\\[-4pt]{\scriptsize\textcolor{ash}{[0.449, 0.488]}}} \\
& MLP & \makecell[t]{63.71\\[-4pt]{\scriptsize\textcolor{ash}{[61.09, 65.98]}}} & \makecell[t]{0.694\\[-4pt]{\scriptsize\textcolor{ash}{[0.664, 0.723]}}} & \textbf{\makecell[t]{63.57\\[-4pt]{\scriptsize\textcolor{ash}{[61.14, 65.84]}}}} & \textbf{\makecell[t]{0.183\\[-4pt]{\scriptsize\textcolor{ash}{[0.177, 0.191]}}}} & \makecell[t]{0.452\\[-4pt]{\scriptsize\textcolor{ash}{[0.432, 0.470]}}} \\
& TabPFN & \makecell[t]{64.42\\[-4pt]{\scriptsize\textcolor{ash}{[61.95, 67.08]}}} & \makecell[t]{0.709\\[-4pt]{\scriptsize\textcolor{ash}{[0.682, 0.739]}}} & \makecell[t]{62.52\\[-4pt]{\scriptsize\textcolor{ash}{[60.21, 64.95]}}} & \makecell[t]{0.187\\[-4pt]{\scriptsize\textcolor{ash}{[0.179, 0.196]}}} & \makecell[t]{0.488\\[-4pt]{\scriptsize\textcolor{ash}{[0.468, 0.507]}}} \\

\end{longtable}

\subsection{Overall Model Performance}

Table~\ref{tab:model_performance} presents the overall performance of the eleven machine learning models using the KNN permutation importance-selected predictor set. These included ten conventional machine learning models and the pretrained tabular foundation model, TabPFN. In the development hold-out test dataset derived from BDHS 2007, 2011, and 2014, balanced accuracy ranged from 64.34\% for KNN to 67.58\% for TabPFN. Among the conventional models, AdaBoost showed the highest observed balanced accuracy at 67.51\%, followed by Random Forest at 67.08\%, Gradient Boosting at 67.06\%, Extra Trees at 66.89\%, and XGBoost at 66.75\%. TabPFN had the highest observed balanced accuracy among all evaluated models in the development hold-out test dataset, at 67.58\%.

AUROC values were similar across several models, with relatively higher balanced accuracy observed for these models in the development hold-out test dataset. TabPFN, AdaBoost, Random Forest, SVM, Gradient Boosting, and XGBoost had AUROC values of approximately 0.73. F1-score was highest for AdaBoost at 66.01\%, followed by TabPFN at 65.98\%, Gradient Boosting at 65.53\%, MLP at 65.53\%, and Random Forest at 65.40\%. Calibration performance varied across models. The lowest Brier score was observed for XGBoost, MLP, and TabPFN, all at 0.212, while the lowest expected calibration error was observed for AdaBoost at 0.217.

\subsection{Model Temporal Robustness}

On the BDHS 2018 test set, balanced accuracy ranged from 63.71\% for MLP to 65.56\% for Gradient Boosting. Gradient Boosting showed the highest observed balanced accuracy in BDHS 2018, followed by Random Forest, XGBoost, AdaBoost, and TabPFN. AUROC values were generally between 0.70 and 0.71, with TabPFN, SVM, Random Forest, Gradient Boosting, XGBoost, and AdaBoost showing similar discrimination. On the BDHS 2022 test set, balanced accuracy ranged from 62.98\% for Logistic Regression to 64.63\% for XGBoost. XGBoost showed the highest observed balanced accuracy in 2022, followed by TabPFN, AdaBoost, Decision Tree, and Gradient Boosting. AUROC values were again similar across several models, generally ranging from 0.69 to 0.71. MLP showed the highest observed F1-score in 2022 and also produced the lowest Brier score and expected calibration error.

\begin{figure}[!t]
     \centering
     \begin{subfigure}[b]{0.49\textwidth}
         \centering
         \includegraphics[width=\textwidth]{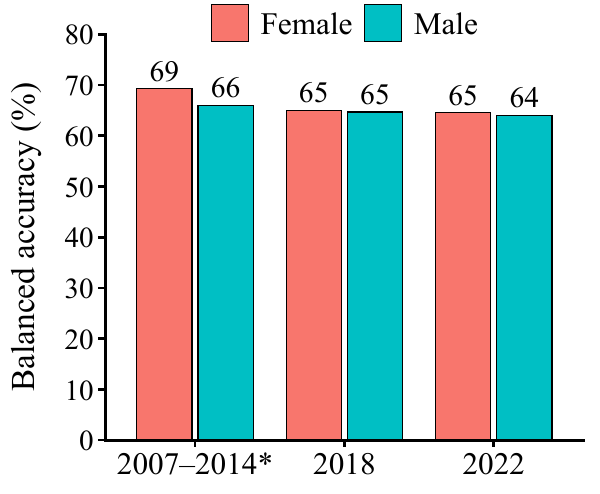}
         \caption{}
         \label{fig:sex}
     \end{subfigure}
     \hfill
     \begin{subfigure}[b]{0.49\textwidth}
         \centering
         \includegraphics[width=\textwidth]{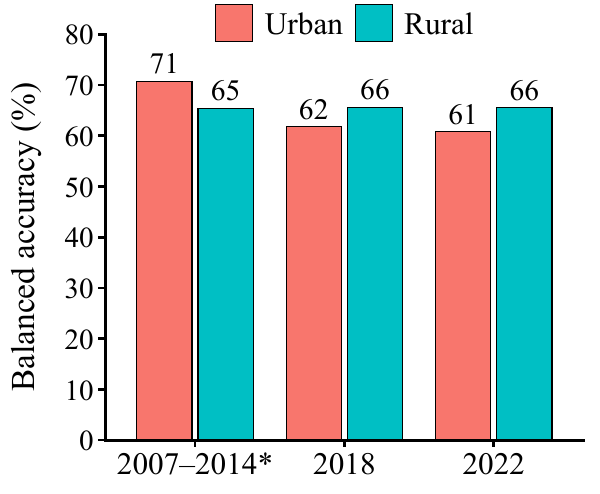}
         \caption{}
         \label{fig:residence}
     \end{subfigure}
     \hfill
     \begin{subfigure}[b]{.75\textwidth}
         \centering
         \includegraphics[width=\textwidth]{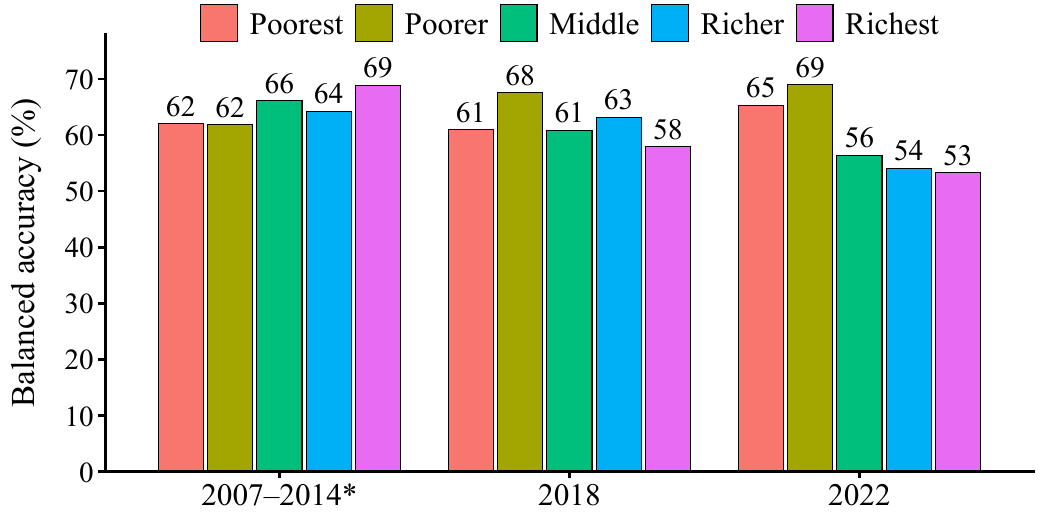}
         \caption{}
         \label{fig:ses}
     \end{subfigure}
        \caption{Balanced accuracy of the TabPFN model across evaluation datasets, stratified by (a) child sex, (b) place of residence, and (c) household socioeconomic status. *The 2007–2014 dataset represents the 20\% development hold-out test set derived from the pooled 2007, 2011, and 2014 surveys; 2018 and 2022 represent the temporal test datasets.}
        \label{fig:fairness}
\end{figure}
\subsection{Model Fairness}

Fig.~\ref{fig:fairness} presents the subgroup balanced accuracy across the development hold-out and temporal test datasets as a fairness assessment. For child sex (Fig.~\ref{fig:sex}), balanced accuracy was slightly higher among female children than male children in the development hold-out test dataset. In the temporal test datasets, the difference by child sex was smaller, although performance remained marginally higher among female children. For place of residence (Fig.~\ref{fig:residence}), balanced accuracy was higher among urban children than rural children in the development hold-out test dataset. However, this pattern changed in the temporal test datasets, where performance was higher among rural children than urban children in both BDHS 2018 and BDHS 2022. This indicates that subgroup performance by residence varied across survey rounds. Balanced accuracy also varied across socioeconomic status groups (Fig.~\ref{fig:ses}). In the development hold-out test dataset, performance was lower among children from the poorest and poorer households and higher among children from the middle and richest households. In the temporal test datasets, the pattern differed. The poorer group showed the highest observed balanced accuracy in both BDHS 2018 and BDHS 2022, whereas performance was lower among the richer and richest groups in BDHS 2022. These findings indicate variation in model performance across socioeconomic groups and support the importance of assessing subgroup performance when evaluating prediction models across population groups.



\begin{figure}[!t]
\centering

\begin{subfigure}[t]{0.32\textwidth}
    \centering
    \includegraphics[width=\textwidth]{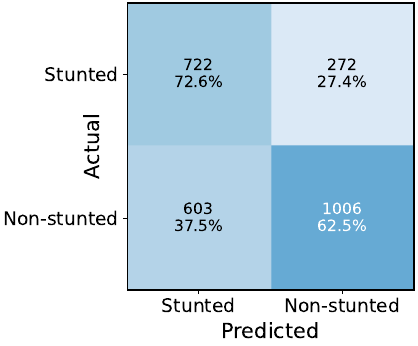}
    \caption{}
\end{subfigure}
\hfill
\begin{subfigure}[t]{0.32\textwidth}
    \centering
    \includegraphics[width=\textwidth]{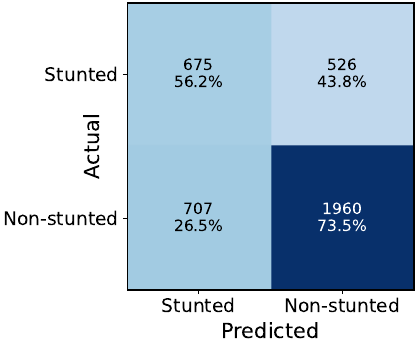}
    \caption{}
\end{subfigure}
\hfill
\begin{subfigure}[t]{0.32\textwidth}
    \centering
    \includegraphics[width=\textwidth]{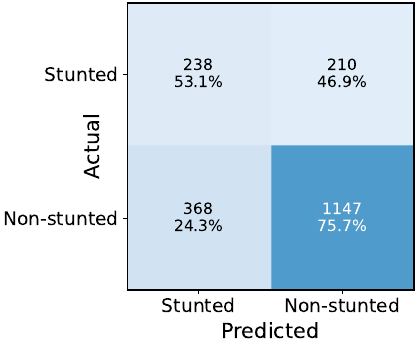}
    \caption{}
\end{subfigure}

\caption{Confusion matrices for TabPFN across the development hold-out and temporal test datasets: (a) 20\% hold-out test dataset derived from BDHS 2007--2014; (b) BDHS 2018; and (c) BDHS 2022.}
\label{fig:confusion}
\end{figure}

Fig.~\ref{fig:confusion} presents the confusion matrices for TabPFN across the three test datasets. In the development hold-out test dataset derived from BDHS 2007, 2011, and 2014, TabPFN correctly classified 72.6\% of stunted children and 62.5\% of non-stunted children. In the BDHS 2018 temporal test dataset, the correct classification of stunted children declined to 56.2\%, whereas the correct classification of non-stunted children increased to 73.5\%. In the BDHS 2022 temporal test dataset, TabPFN correctly classified 53.1\% of stunted children and 75.7\% of non-stunted children. These findings show a shift in classification performance across survey periods, with higher correct classification of stunted children in the development hold-out test dataset and higher correct classification of non-stunted children in the temporal test datasets.

\section{Discussion}

\paragraph{Principal findings.} This study used repeated nationally representative BDHS data to predict childhood stunting and identify prediction-relevant predictors. The models showed moderate predictive performance, but performance varied across datasets, metrics, and population subgroups. In the development hold-out test dataset derived from BDHS 2007, 2011, and 2014, TabPFN and AdaBoost showed the highest observed balanced accuracy. In the temporal test datasets, Gradient Boosting showed the highest observed balanced accuracy in BDHS 2018, while XGBoost showed the highest observed balanced accuracy in BDHS 2022. These findings indicate that no single model consistently showed the highest observed performance across all evaluation settings.

Previous machine learning studies in Bangladesh and other low- and middle-income countries have reported that ensemble models, including Random Forest, Gradient Boosting, and XGBoost, often perform well for predicting childhood undernutrition \cite{anku2024,bitew2022,chilyabanyama2022,fenta2021,islam2024,mansur2021,rahman2021,shen2023,talukder2020,tamanna2025}. The present study showed a similar pattern, with AdaBoost, Gradient Boosting, XGBoost, Random Forest, and TabPFN showing relatively strong performance. However, direct numerical comparison with earlier studies is difficult because previous studies used different datasets, outcome definitions, predictor sets, validation strategies, and performance metrics. Many earlier studies relied on a single survey round and within-sample validation, whereas this study evaluated model performance using a development hold-out dataset and temporally distinct test datasets. Therefore, this study extends previous work by examining how model performance changes across survey periods and population subgroups, rather than focusing only on performance within a single dataset.

\paragraph{Prediction-relevant predictors.} The selected predictors covered several epidemiological domains, including child age and sex, breastfeeding, recent fever, vitamin A supplementation, maternal height, maternal age at first birth, antenatal care, delivery mode, household socioeconomic status, sanitation, father's education and occupation, and mass media exposure. These variables are consistent with previous studies showing that childhood stunting in Bangladesh is associated with child, maternal, household, socioeconomic, and health-service factors \cite{alom2012,chowdhury2016}. In the present study, these variables should be interpreted as prediction-relevant predictors rather than causal determinants. Their selection reflects their contribution to prediction within the study data and does not imply that they independently cause childhood stunting.

\paragraph{Temporal and subgroup performance:} Model performance generally decreased in the temporal test datasets compared with the development hold-out test dataset, although the magnitude and direction of change varied across models and performance metrics. This suggests that prediction models developed from earlier survey rounds may not perform equally well in later survey periods. This finding is important because previous machine learning studies on childhood malnutrition in Bangladesh have mostly relied on single datasets and within-sample validation \cite{islam2024,tamanna2025}. Stratified analyses showed that model performance varied by child sex, residence, and socioeconomic status. Performance was generally higher among female and urban children than among male and rural children in the development hold-out test dataset, although this pattern was less consistent in the temporal test datasets. Model performance also differed across socioeconomic groups. These findings suggest that overall model performance alone may mask differences in performance across population subgroups. Therefore, subgroup performance should be considered when evaluating prediction models for potential use in public health planning.

\paragraph{Limitations.} This study has several limitations. First, a complete-case approach was used, and children with missing anthropometric or predictor information were excluded. Some anthropometric and maternal health-care variables were not collected for all participants in the original BDHS surveys. Missing stunting outcomes were not imputed or estimated because the outcome could not be determined without the required anthropometric measurements. This reduced the final analytic sample and may limit generalizability. Second, the models were restricted to variables available and comparable across BDHS rounds, excluding potentially relevant contextual factors such as food security, climate, and local environment. Third, although random undersampling was performed only within training folds and not on validation or test datasets, the difference in class distribution between training and evaluation data may have affected probability calibration. Finally, feature selection and feature importance indicate prediction relevance rather than causality.

\paragraph{Future Directions.} Future studies could improve generalizability by integrating additional contextual and geospatial data, such as climate, agriculture, food security, and environmental indicators. External validation using MICS or DHS datasets from other settings would help assess whether the models generalize beyond the current BDHS data. Further work could also evaluate calibration and additional fairness measures, develop age-specific models, and use model interpretation methods such as SHAP or partial dependence plots to better understand how prediction-relevant predictors contribute to stunting prediction.

\section{Conclusion}

This study developed and evaluated eleven machine learning models for predicting childhood stunting in Bangladesh using repeated nationally representative BDHS data. The models showed moderate predictive performance, but performance varied across datasets, metrics, and population subgroups. In the development hold-out test dataset, TabPFN and AdaBoost showed the highest observed balanced accuracy, while Gradient Boosting and XGBoost showed the highest observed balanced accuracy in the BDHS 2018 and BDHS 2022 temporal test datasets, respectively. These patterns indicate that no single model consistently showed the highest observed performance across all evaluation settings. The findings highlight the importance of temporal evaluation and subgroup performance assessment in public health prediction modeling. Model performance varied across survey rounds, suggesting that models developed using earlier BDHS data may not perform equally well in later survey periods. Subgroup analyses also showed variation in balanced accuracy by child sex, place of residence, and socioeconomic status, indicating that overall performance metrics may mask important differences across population groups.


%
\bibliographystyle{splncs04}
\bibliography{AusDM}
\end{document}